\documentclass[conference]{IEEEtran}
\IEEEoverridecommandlockouts                              % This command is only
\usepackage[utf8]{inputenc}
\usepackage[T1]{fontenc}
\usepackage{amsmath} 
\usepackage{booktabs}
\usepackage{multirow} 
\usepackage{graphicx}
\usepackage{hyphenat} 
\usepackage{url} 
\title{\LARGE \bf
Who Benefits From Sinus Surgery? Comparing Generative AI and Supervised Machine Learning for Predicting Surgical Outcomes in Chronic Rhinosinusitis
}

\author{Sayeed Shafayet Chowdhury$^{1,2}$, Snehasis Mukhopadhyay$^{1}$, Shiaofen Fang$^{2}$, and Vijay R. Ramakrishnan$^{3}$% <-this % stops a space
\thanks{$^{1}$Sayeed Shafayet Chowdhury (corresponding author) and Snehasis Mukhopadhyay are with the Department of Computer Science,
        Purdue University, Indianapolis, IN, USA {\tt\small chowdh23@purdue.edu, saychow@iu.edu}}
\thanks{$^{2}$Shiaofen Fang is with the Department of Computer Science, Indiana University Indianapolis, IN, USA}
\thanks{$^{3}$Vijay R. Ramakrishnan is with the Department of Otolaryngology—Head and Neck Surgery, Indiana University School of Medicine, Indianapolis, IN, USA
        {\tt\small vrramakr@iu.edu}}%
}

\begin{document}

\maketitle
\thispagestyle{empty}
\pagestyle{empty}

%%%%%%%%%%%%%%%%%%%%%%%%%%%%%%%%%%%%%%%%%%%%%%%%%%%%%%%%%%%%%%%%%%%%%%%%%%%%%%%%
\begin{abstract}

Artificial intelligence has reshaped medical imaging, yet the use of AI on clinical data for prospective decision support remains limited. We study pre-operative prediction of clinically meaningful improvement in chronic rhinosinusitis (CRS), defining success as a $\geq$ 8.9-point reduction in SNOT-22 at 6 months (MCID). In a prospectively collected cohort where \textit{all} patients underwent surgery, we ask whether models using only pre-operative clinical data could have identified those who would have poor outcomes, i.e. those who should have avoided surgery. We benchmark supervised ML (logistic regression, tree ensembles, and an in-house MLP) against generative AI (ChatGPT, Claude, Gemini, Perplexity), giving each the same structured inputs and constraining outputs to binary recommendations with confidence. Our best ML model (MLP) achieves ~85\% accuracy with superior calibration and decision-curve net benefit; GenAI models underperform on discrimination and calibration across zero-shot setting. Notably, GenAI justifications align with clinician heuristics and the MLP’s feature importance, repeatedly highlighting baseline SNOT-22, CT/endoscopy severity, polyp phenotype, and physchology/pain comorbidities. We provide a reproducible tabular-to-GenAI evaluation protocol and subgroup analyses. Findings support an ML-first, GenAI-augmented workflow: deploy calibrated ML for primary triage of surgical candidacy, with GenAI as an explainer to enhance transparency and shared decision-making.

\end{abstract}


%%%%%%%%%%%%%%%%%%%%%%%%%%%%%%%%%%%%%%%%%%%%%%%%%%%%%%%%%%%%%%%%%%%%%%%%%%%%%%%%
\section{Introduction}

Chronic Rhinosinusitis (CRS) is clinically heterogeneous, and patients may experience variable quality-of-life (QoL) gains following endoscopic sinus surgery (ESS) \cite{c1,c5}. This variability complicates pre-operative decision-making, underscoring the need for \emph{individualized} pre-operative benefit prediction that can inform shared decisions and set expectations. This consideration is particularly salient within value-based care models that are increasingly shaping contemporary healthcare delivery.  We operationalize clinical benefit using the Sino-Nasal Outcome Test-22 (SNOT-22) minimal clinically important difference (MCID), defining success as a $\geq 8.9$-point reduction at six months post-operatively \cite{c2}. Objective disease burden, summarized by the Lund–Mackay CT score and endoscopic severity scales, together with phenotype (CRSwNP vs.\ CRSsNP) and comorbidities further modulate expected benefit observed in cohort studies of ESS outcomes \cite{c3,c4,c5}.

Large language models (LLMs) are increasingly used for clinical reasoning \cite{c7}, yet their performance relative to tuned, supervised \emph{tabular} machine-learning (ML) models trained on structured pre-operative data remains unclear. While ML on electronic health records has shown promise \cite{c6}, rigorous, head-to-head comparisons between calibrated tabular models and multiple generative AI (GenAI) systems using the same inputs, clinically grounded endpoints, and responsible-use constraints are sparse. Addressing this gap, we construct a transparent benchmark to test whether pre-operative data can be used to identify patients unlikely to achieve MCID (i.e., those who, in retrospect, should not have been recommended surgery), and to quantify trade-offs between ML classifiers and several GenAI systems under standardized prompting.

The main contributions of this work are- (1) To our knowledge, this is the \emph{first} comprehensive evaluation of \textbf{generative AI (GenAI)} systems for \emph{predictive clinical outcome modeling} on \textbf{structured, pre-operative} data in chronic rhinosinusitis (CRS). We benchmark multiple state-of-the-art LLMs (e.g., ChatGPT, Claude, Gemini, Perplexity) against tuned, calibrated tabular ML baselines (logistic regression, tree ensembles, and an in-house MLP) on the clinically grounded endpoint of achieving SNOT-22 MCID. (2) We provide standardized expert-guided zero-shot prompts that serialize structured inputs, constrain outputs to binary recommendations with confidence, and analyze the results. (3) We also examine the reasoning process of GPT models \cite{c7,c8} and, notably, find that the features they emphasize align closely with expert judgment and with the top predictors identified by our ML model.
 (4) This framework enables practical safeguards for real-world deployment (PHI handling, model/version transparency) and an \emph{ML-first, GenAI-augmented} workflow in which a calibrated classifier provides primary triage and GenAI serves as a secondary explainer to enhance transparency and shared decision-making \cite{c6,c7}. Together, these elements establish a clinically grounded, reproducible comparison of ML and GenAI for pre-operative CRS decision support, emphasizing not only accuracy but also calibration, net benefit, and responsible use.

\section{Background \& Related Work}

\textbf{ESS outcome prediction with supervised ML.}  Numerous studies have studied endoscopic sinus surgery (ESS) outcomes using pre-operative clinical and imaging summaries to identify risk factors for failure, with emphasis on patient-reported symptoms (e.g., SNOT-22), objective disease burden (CT/endoscopy), phenotype (CRSwNP vs.\ CRSsNP), and key comorbidities. Multi-institutional cohort work has established that baseline QoL and objective disease severity are associated with post-operative gains, while mood/pain comorbidities can attenuate benefit \cite{c5}. These variables map naturally to tabular machine-learning (ML) pipelines (logistic regression, tree ensembles, neural networks) and provide a structured substrate for prediction once leakage is controlled and calibration is addressed. Our use of SNOT-22 as the primary endpoint and CT/endoscopy as objective correlates follows standard practice in CRS outcomes research \cite{c2,c3,c4,c5,rudmik2015patient}.

\textbf{LLMs for clinical reasoning and tabular inference.}  Large language models (LLMs) have demonstrated broad clinical knowledge and emergent reasoning capabilities on textual tasks \cite{c7}, and can produce confidence-like signals that partially reflect uncertainty awareness \cite{c8}. However, when repurposed for \emph{predictive} inference on \emph{structured} clinical data, important limitations arise: (i) sensitivity to prompt phrasing and few-shot selection, (ii) instability across sampling runs, and (iii) weaker probability calibration than tuned tabular ML models. Consequently, rigorous comparison against calibrated ML baselines is essential, using metrics that reflect both discrimination and clinical decision quality (e.g., F1 score, AUC) \cite{c10,c9}.

\textbf{CRS guidelines and MCID rationale.}  Contemporary CRS guidance emphasizes symptom severity and objective disease burden in surgical selection, with phenotype-stratified considerations and shared decision-making \cite{c1,aao_indicators_ess,aao_guideline_adultsinusitis_2015,icar_rs_2021}. The SNOT-22 is a validated instrument \cite{c2}; an improvement of $\geq 8.9$ points at follow-up is widely used as a pragmatic minimal clinically important difference (MCID) to denote meaningful benefit at the patient level. Objective disease is commonly summarized by the Lund–Mackay CT score and endoscopic severity scales (e.g., Lund–Kennedy), which correlate with anatomic disease and, in aggregate, with expected post-operative gains \cite{c3,c4,c5}.

\textbf{Positioning of this study.}  Prior LLM evaluations in healthcare have largely focused on question answering, summarization, or qualitative reasoning \cite{c7,c8}, while prior ESS outcome studies have relied on supervised ML within single method families and without GenAI comparators \cite{c5}. Most of these studies are about making diagnosis of complex cases, not guiding or predicting trajectory. Our study differs in the following ways: (i) focus is on predicting surgery outcome and trajectory, %\emph{same dataset, same endpoint, identical evaluation} across all models to ensure a fair head-to-head comparison; (
ii) performance analysis of \emph{multi-model GenAI} benchmarks (ChatGPT, Claude, Gemini, Perplexity) under standardized input prompts and constrained outputs; (iii) strong emphasis on \emph{calibration and clinical utility} (F1 score, AUC, P-R curves) alongside discrimination \cite{c9,c10}, (iv) investigating the feature importance as perceived by our in-house ML model as well as the decision logic of the GenAI models and their alignment with human experts.

\section{Dataset}

To train predictors of surgical benefit for CRS, we focus on pre-operative, routinely available clinical factors and a patient-centered outcome. The Sino-Nasal Outcome Test-22 (SNOT-22) is the prevailing QoL instrument for CRS, comprising 22 items rated 0–5 (total 0–110; higher indicates worse QoL) and extensively validated for symptom severity and change detection \cite{Le2018,Soler2018Laryngoscope,Mattos2019}.  

We leverage data from an NIH-funded, multicenter CRS outcomes study (clinicaltrials.gov: NCT01332136) that prospectively captured demographics, comorbidities, phenotype, imaging/endoscopy summaries, and patient-reported outcomes, including SNOT-22, a general health utility (SF-6D–derived HUV) \cite{Brazier2002}, and olfactory measures using a standardized $\sim$30-field pre-operative feature set accessible in routine practice. Two cohorts were available: a set of 791 patients (50 recorded attributes; expected missingness in some fields; “2R01”) and a separate set of 355 patients with comparable clinician-recorded metadata (“3R01”). For this study’s objective: \emph{identifying surgical candidates unlikely to achieve meaningful improvement}, we merged these 2 sets and restricted analysis to patients who actually underwent endoscopic sinus surgery (ESS), excluding those managed medically. After concatenation and filtering to ESS only, the combined analytic dataset comprised \textbf{524} surgical cases.

\section{Methods}

\subsection{Pre-processing \& Feature Construction}
From the dataset, we removed post-operative fields (e.g., 6-month SNOT-22, post-op HUV, follow-up olfaction) to prevent leakage, and harmonized variable names across cohorts. We retained routinely available pre-op features (demographics, socioeconomic factors, phenotype, comorbidities, CT/endoscopy scores, baseline SNOT-22). Categorical variables were deterministically encoded via fixed dictionaries (e.g., \texttt{SEX}: Female$\rightarrow$0, Male$\rightarrow$1; \texttt{INSURANCE}: ordered categories); binary history/comorbidity fields were mapped to $\{0,1\}$; continuous measures (SNOT-22 baseline, CT Lund–Mackay, endoscopy) were kept numeric. Obvious placeholders (e.g., “None”, blanks) were normalized; remaining missingness was handled by removing the data. We performed strict leakage checks: (i) no post-op variables in features, (ii) all preprocessing state (encoders, imputers, scalers) fit within CV folds, (iii) no patient overlap across splits. Finally, we concatenated the two cleaned cohorts and executed a stratified 80/20 train–test split on the binary target (MCID $\geq$ 9), preserving class prevalence. Class imbalance was addressed via class weights (for LR/trees) and focal/weighted loss (for MLP). All random seeds and preprocessing artifacts were versioned for reproducibility.

\subsection{Supervised ML Models}

We evaluated five classifiers on the held-out test set using identical pre-processing, and class weighting. The per-class precision/recall and overall accuracy are summarized in Table~\ref{tab:per_class_metrics}. 
The MLP provides the strongest overall profile (high accuracy, well-calibrated probabilities, and the best minority-class recall among competitive models). Therefore, in subsequent analyses and comparisons against GenAI systems, we use the MLP as the primary supervised baseline. In this case, the MLP included 1 hidden layer with 400 neurons as the architecture.

\begin{table}[t]
\centering
\footnotesize
\caption{Per-class precision/recall and overall accuracy on the held-out test set.}
\label{tab:per_class_metrics}
\begin{tabular}{lcccccc}
\toprule
\multirow{2}{*}{\textbf{Model}} & \multirow{2}{*}{\textbf{Acc}} & \multicolumn{2}{c}{\textbf{Class 0}} & \multicolumn{2}{c}{\textbf{Class 1}} \\
\cmidrule(lr){3-4} \cmidrule(lr){5-6}
 &  & \textbf{Prec} & \textbf{Rec} & \textbf{Prec} & \textbf{Rec} \\
\midrule
Logistic Regression & \textbf{0.85} & 0.75 & 0.30 & 0.86 & 0.98 \\
SVM                 & 0.79          & 0.33 & 0.10 & 0.82 & 0.95 \\
Na\"ive Bayes       & 0.30          & 0.21 & 1.00 & 1.00 & 0.13 \\
MLP                 & \textbf{0.85} & \textbf{0.64} & \textbf{0.45} & \textbf{0.88} & \textbf{0.94} \\
Random Forest       & 0.82          & 0.56 & 0.25 & 0.84 & 0.95 \\
\bottomrule
\end{tabular}

\begin{minipage}{\columnwidth}
\footnotesize
\emph{Notes:} Class~0 = “surgery does not yield desired QoL improvement” (MCID $<9$); Class~1 = “surgery results in desired QoL improvement" (MCID $\ge 9$).
\end{minipage}
\end{table}

\begin{figure}[t]
  \centering
  \includegraphics[width=\columnwidth]{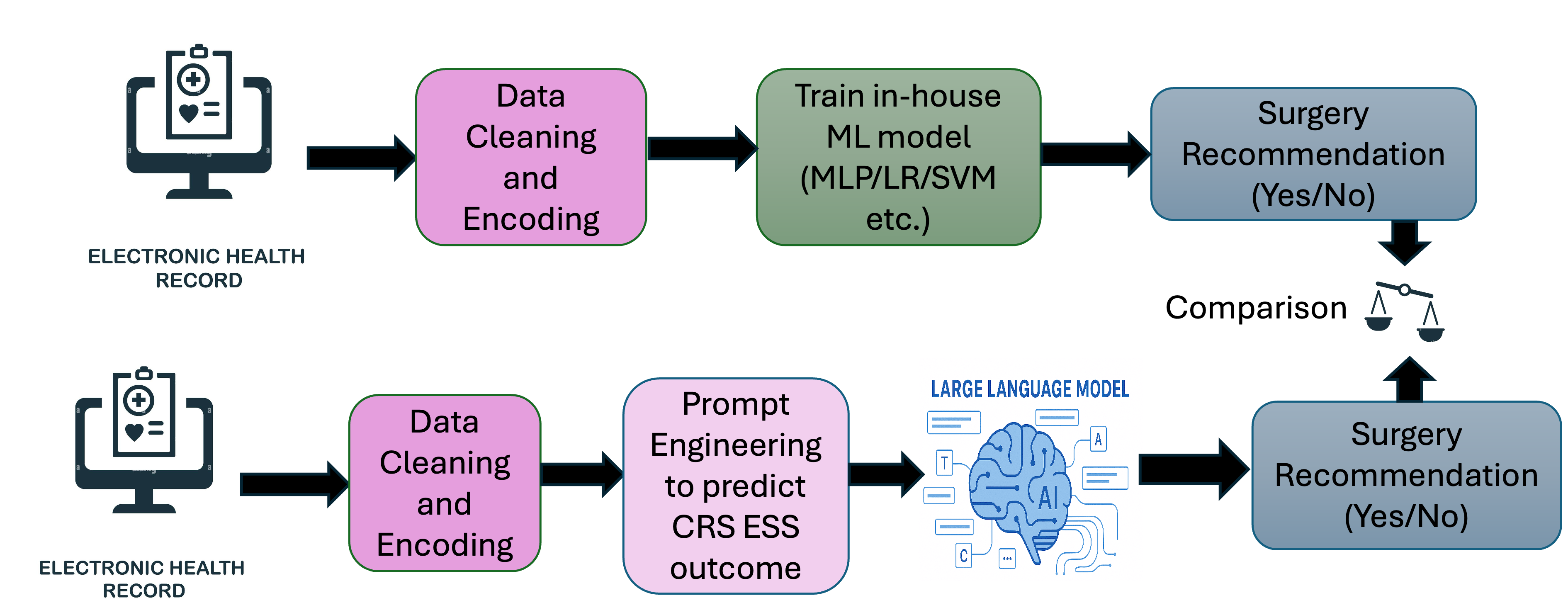}
  \caption{Study pipeline. Pre-operative EHR variables are cleaned/encoded and used in two parallel paths: (top) calibrated supervised ML (LR/RF/MLP) and (bottom) tabular-to-LLM prompting for GenAI (ChatGPT/Claude/Gemini/Perplexity). Both output binary surgery recommendations (MCID $\geq$ 9 target).}
  \label{fig:schematic}
\end{figure}

% \paragraph{Input serialization.}
% Each patient is converted to a \textbf{deterministic, order-fixed} key--value schema comprising only pre-operative fields (demographics, phenotype, comorbidities, baseline SNOT-22, CT Lund--Mackay, endoscopy score, etc.). Categorical features are rendered with canonical labels; missing values are serialized as \texttt{NA}. Example (truncated):
% \begin{verbatim}
% AGE: 58; SEX: Male; POLYPS: No; SMOKER: No;
% DEPRESSION: Yes; FIBROMYALGIA: No;
% SNOT22_BASELINE: 51; CT_LUND_MACKAY: 8;
% ENDOSCOPY_SCORE: 4; PREV_SURGERY: No; ...
% \end{verbatim}

\subsection{GenAI Inference Protocol}

\paragraph{Models and versioning.}
We benchmark five LLMs available on 2025--10--21: \emph{ChatGPT (GPT-5 Thinking)}, \emph{MedGPT}, \emph{Gemini~2.5~Pro}, \emph{Perplexity Sonar}, and \emph{Claude Sonnet~4.5}. For each run we log vendor, model identifier, and access date, e.g.,
\texttt{gpt-5-thinking (2025--10--21)}, \texttt{medgpt (2025--10--21)}, \texttt{gemini-2.5-pro (2025--10--21)}, \texttt{perplexity sonar (2025--10--21)}, \texttt{claude-sonnet-4.5 (2025--10--21)}.

\paragraph{Prompting conditions.}
% We evaluate two settings:
% \begin{itemize}
%   \item \textbf{Zero-shot clinical reasoning:} concise task description, MCID threshold (8.9), strict output schema; \emph{no} chain-of-thought requested.
%   \item \textbf{Few-shot:} \textbf{3--5} templated exemplars from the \emph{training} split only (never from test), showing serialized pre-op inputs and required outputs; exemplars span typical and edge cases.
% \end{itemize}

We evaluate the following:
\begin{itemize}
  \item \textbf{Zero-shot clinical reasoning:} concise task description, MCID threshold (8.9), strict output schema; chain-of-thought requested, as we asked the model to explain its decision-making.
  % \item \textbf{Few-shot:} \textbf{3--5} templated exemplars from the \emph{training} split only (never from test), showing serialized pre-op inputs and required outputs; exemplars span typical and edge cases.
\end{itemize}

\paragraph{Constrained outputs.}
To ensure comparability, the prompt enforces:
\begin{itemize}
  \item \textbf{PREDICTION:} \texttt{0} or \texttt{1}.
  \item \textbf{CONFIDENCE:} one of \{\emph{very confident}, \emph{Somewhat confident}, \emph{Neutral}, \emph{Somewhat unsure}, \emph{Not at all confident}\}.
\end{itemize}
We parse \texttt{PREDICTION} as the class label. 

\paragraph{Sampling and stability.}
Each case is run for $k{=}5$ replicates with fixed decoding parameters (temperature $t$, top-$p$, max tokens) recorded per model; default ranges: $t\in[0.1,0.5]$, top-$p\in[0.7,0.95]$. The final GenAI prediction uses \emph{majority vote} over the $k$ outputs; ties are broken by the mean proxy score. We report replicate variability and fix random seeds where supported.

\paragraph{Safety, governance, and logging.}
Inputs contain only de-identified, structured \emph{pre-operative} variables; post-operative fields are excluded to prevent leakage. We log model name/version/date, prompt hash, decoding parameters, raw outputs, parsed labels, and parser status, enabling audit and exact re-runs.

\paragraph{Canonical prompt used.}
{\small
\begin{quote}
\textbf{Prompt:} Assume you are an expert Otolaryngologist with a special interest in Rhinology and endoscopic sinus surgery. I will provide you an excel sheet with different patients' clinical information. They have been suffering from CRS (chronic rhinosinusitis), and have undergone prior appropriate medical therapies, and are considering whether or not to have endoscopic sinus surgery. In this discussion, we have the baseline sinonasal quality of life measured by the SNOT22 score, and you need to predict what their result might be at 6-months postoperatively, should they choose to undergo surgery. Consider a decrease in total SNOT22 of more than 8.9 as clinically significant, and the surgery would be deemed successful. Based on the given data, provide your predictions in the form of 0 or 1. 0 means the patient would not be expected to achieve an 8.9 point improvement in SNOT22 and the surgery should not be recommended; 1 means the patient is expected to achieve greater than an 8.9 point improvement in total SNOT22 and the surgery should be recommended. Also complete the confidence column, your options would be \emph{very confident}, \emph{Somewhat confident}, \emph{Neutral}, \emph{Somewhat unsure}, \emph{Not at all confident}. The data are in the uploaded csv file. All patients meet criteria to have surgery by current clinical guidelines, but we have observed from previous published data and our own experience that some patients will not achieve their expected outcome. Your job is to predict whether surgery should be recommended for these patients, or not.
\end{quote}
}

\paragraph{Evaluation.}
GenAI outputs are compared to supervised ML on the identical test split and endpoint. We report AUROC/AUPRC, threshold metrics (F1, sensitivity, specificity), calibration (Brier score, reliability curves), and decision-curve \emph{net benefit}, with paired tests (DeLong, bootstrap CIs, McNemar) for head-to-head comparisons.

\begin{figure}[t]
  \centering
  \includegraphics[width=0.74\columnwidth]{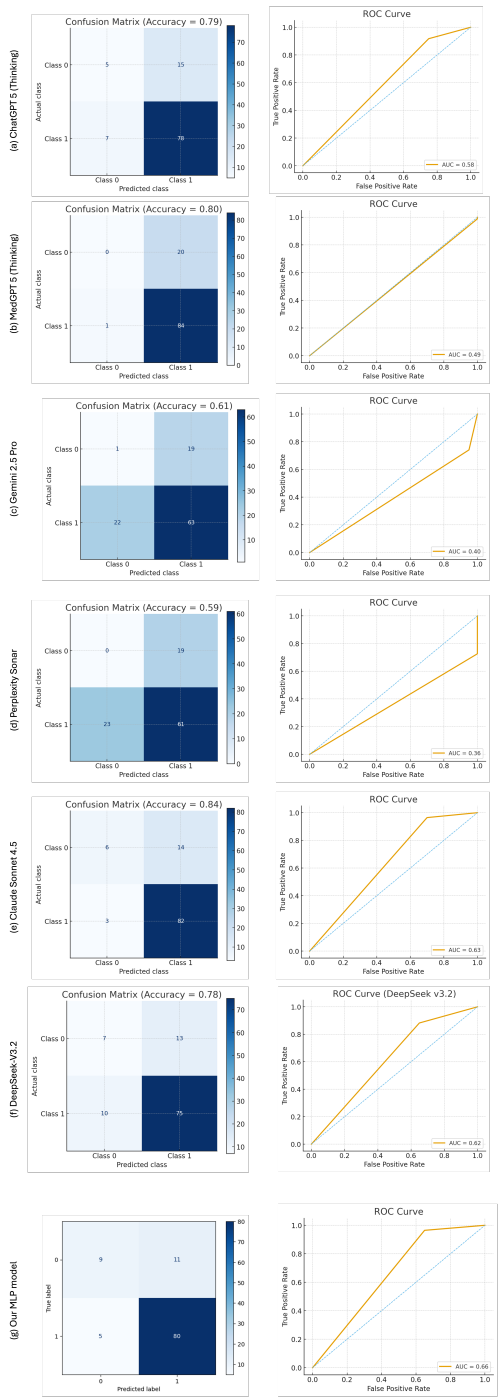}
  \caption{Confusion matrices (left of each pair) and ROC curves (right) for six models on the CRS test set ($n{=}105$). Rows (top$\to$bottom): ChatGPT 5 (Thinking), MedGPT\textendash 5 (thinking), Gemini 2.5 Pro, Perplexity Sonar, Claude Sonnet 4.5, and our MLP. Numeric overlays show counts; ROC panels report the AUC for each model.}
  \label{fig:big_results_grid}
\end{figure}

\section{Results}

\subsection{ChatGPT 5 (Thinking)}
As shown in Fig.~\ref{fig:big_results_grid}(a), ChatGPT~5 (Thinking) attains accuracy $0.79$ and ROC--AUC $0.58$ with confusion matrix $[5,15;\,7,78]$. Positive-class performance is good (Class~1 precision $\approx 0.84$, recall $\approx 0.92$), but detection of non-responders is limited (Class~0 precision $\approx 0.42$, recall $=5/20=0.25$). The error profile is dominated by Class~0$\rightarrow$1 flips (15 false positives), indicating a bias toward recommending surgery relative to the supervised baseline.

\subsection{MedGPT\textendash 5 (Thinking)}
As demonstrated in Fig.~\ref{fig:big_results_grid}(b), MedGPT\textendash 5 attains accuracy $0.80$ and ROC--AUC $0.49$ with confusion matrix $[0,20;\,1,84]$. The model is strongly biased toward recommending surgery (Class~1): Class~1 recall is $84/85\approx0.99$ with precision $84/(84+20)\approx0.81$, whereas Class~0 recall is $0/20=0$ and precision $0/(0+1)=0$. This yields twenty false positives and no true negatives, producing near-random discrimination despite apparent high accuracy, and offering little utility for sparing likely non-responders.

\subsection{Gemini 2.5 Pro}
Fig.~\ref{fig:big_results_grid}(c) depicts results for Gemini~2.5~Pro which attains accuracy $0.61$ and ROC--AUC $0.40$ with confusion matrix $[1,19;\,22,63]$. Positive-class performance is moderate (Class~1 precision $\approx 63/(63+19)=0.77$, recall $63/85\approx 0.74$), whereas detection of non-responders is very limited (Class~0 precision $\approx 1/(1+22)=0.04$, recall $1/20=0.05$). Errors are dominated by Class~0$\rightarrow$1 flips (19 false positives), reducing discrimination and clinical utility relative to the supervised baseline.

\subsection{Perplexity Sonar}
Fig.~\ref{fig:big_results_grid}(d) shows results for the Perplexity Sonar model which attains accuracy $0.59$ and ROC--AUC $0.36$ with confusion matrix $[0,19;\,23,61]$. Positive-class performance is moderate (Class~1 precision $61/(61+19)\approx0.76$, recall $61/(61+23)\approx0.73$), but the model fails to identify any non-responders (Class~0 precision $=0$, recall $=0$), producing 19 false positives and no true negatives. This treat-all tendency substantially limits discrimination and clinical usefulness relative to the supervised baseline.

\subsection{Claude Sonnet 4.5}
Fig.~\ref{fig:big_results_grid}(e) illustrates Claude Sonnet~4.5 results, which achieves accuracy $0.84$ and ROC--AUC $0.63$ with confusion matrix $[6,14;\,3,82]$. Positive-class performance is strong (Class~1 precision $82/(82+14)\approx0.85$, recall $82/85\approx0.97$), while detection of non-responders remains modest (Class~0 precision $6/(6+3)=0.67$, recall $6/20=0.30$). The error profile is dominated by Class~0$\rightarrow$1 flips (14 false positives), indicating a bias toward recommending surgery, though overall discrimination is meaningfully better than other GenAI models.

\subsection{DeepSeek-V3.2}
Fig.~\ref{fig:big_results_grid}(f) presents the performance of the DeepSeek-V3.2 model, which attains an accuracy of $0.78$ and a ROC--AUC of $0.62$, with confusion matrix $[7,13;,10,75]$. The model demonstrates robust positive-class identification (Class~1 precision $75/(75+13)\approx0.85$, recall $75/85\approx0.88$), effectively capturing most surgery-appropriate cases. However, performance for the negative class remains limited (Class~0 precision $7/(7+10)\approx0.41$, recall $7/20=0.35$). The dominant error mode involves Class~0$\rightarrow$1 misclassifications (13 false positives), again reflecting a tendency to over-recommend surgery. Nevertheless, DeepSeek-V3.2 achieves a stable trade-off between recall and precision, yielding competitive discrimination compared with prior GenAI variants.

\subsection{Multi-Layer Perceptron (In-house top performer)}
The bottom row presents the best overall profile with confusion matrix $[9,11;\,5,80]$ and ROC--AUC $\approx 0.66$. Class~1 precision/recall remain high ($\approx 0.88/0.94$) while Class~0 recall improves to $0.45$, providing the strongest balance for identifying patients unlikely to meet MCID and serving as our primary supervised baseline for subsequent comparisons.

\begin{table}[t]
\centering
\footnotesize
\caption{Comparison across models.}
\label{tab:overall_perf_compact}
\begin{tabular}{lccc}
\toprule
\textbf{Model / Variant} & \textbf{AUROC} & \textbf{F1} & \textbf{AP} \\
\midrule
ChatGPT 5 (Thinking)      & 0.58 & 0.77 & 0.84 \\
MedGPT--5 (Thinking)      & 0.49 & 0.72 & 0.81 \\
Gemini 2.5 Pro            & 0.40 & 0.62 & 0.78 \\
Perplexity Sonar          & 0.36 & 0.61 & 0.78 \\
Claude Sonnet 4.5         & 0.63 & 0.81 & 0.85 \\
DeepSeek-V3.2         & 0.62 & 0.77 & 0.85 \\
\textbf{MLP (ours)}       & \textbf{0.66} & \textbf{0.83} & \textbf{0.86} \\
\bottomrule
\end{tabular}
\end{table}

\subsection{Overall comparison and takeaway}

Table~\ref{tab:overall_perf_compact} summarizes the head–to–head benchmark. Our calibrated \emph{MLP} attains the best overall discrimination (AUROC $=0.66$) and F1 score ($0.83$) while also achieving the highest average precision (AP) among the competitive models (AP$=0.86$ vs.\ $0.85$ for Claude Sonnet~4.5 and DeepSeek-V3.2). In contrast, the GenAI systems exhibit lower AUROC (0.36–0.63) and comparable or lower F1, reflecting weaker separation of responders from non-responders at clinically relevant thresholds.

Crucially, this cohort is \emph{highly imbalanced} - every patient actually underwent surgery, and our aim is to retrospectively identify the Class~0 (minority class) cases, patients who likely would \emph{not} achieve MCID despite guideline eligibility. Under such imbalance, overall accuracy is not a fair metric; a model can appear “good” by recommending surgery for nearly everyone while failing miserably at finding true 0 cases. Consistent with the confusion-matrix analyses, the MLP delivers the strongest minority-class performance (higher Class~0 recall with reasonable precision), whereas several GenAI models collapse toward a treat-all pattern.

Taken together, the calibrated MLP offers the best combination of discrimination (highest AUROC), ranking ability (highest AP among top models), and practical utility for identifying likely non-responders, making it the most suitable primary classifier for our ML-first, GenAI-augmented workflow.

\subsection{Probing ChatGPT’s decision logic}

To understand \emph{how} the LLM arrived at recommendations, we explicitly queried ChatGPT (GPT\textendash 5 Thinking) for its rationale under the same input and constrained outputs used for benchmarking. Without revealing test labels, we asked the model to state the logic it applied to convert pre\hyp operative features into a predicted 6\hyp month SNOT\hyp 22 change and a binary decision (MCID $\geq 8.9$). The model produced a consistent, rule\hyp based schema that can be summarized as a shallow decision tree with multiplicative modifiers:

\paragraph{Base expected improvement.}
A baseline anchor equal to \emph{45\% of the baseline SNOT\hyp 22} was taken as the initial expected drop, reflecting the clinical intuition that higher pre\hyp operative symptom burden allows larger absolute reductions after ESS (commonly 20--30 point medians in high\hyp burden cohorts).

\paragraph{“Room to improve’’ and objective disease.}
The base change was scaled by bracketed factors for baseline SNOT\hyp 22 (\(<25:\times0.5\), 25--39:\(\times0.7\), 40--59:\(\times1.0\), 60--79:\(\times1.1\), \(\ge80:\times1.2\)), endoscopy severity (\(\le3:\times0.8\), 4--6:\(\times0.9\), 7--10:\(\times1.0\), \(\ge11:\times1.1\)), CT Lund--Mackay (\(\le6:\times0.85\), 7--12:\(\times1.0\), \(>12:\times1.1\)), and a CRSwNP bonus (\(\times1.05\)) to reflect greater average gains with polyp debulking.

\paragraph{Negative outcome modifiers.}
The model then applied penalty multipliers when specific comorbidities or histories were present: depression \(\times0.7\), fibromyalgia \(\times0.7\); smoker \(\times0.85\), COPD \(\times0.8\), asthma \(\times0.9\); OSA \(\times0.9\), diabetes \(\times0.9\), GERD \(\times0.95\), ASA intolerance \(\times0.9\); previous surgery \(\times0.85\); age \(\ge65:\times0.9\). These choices mirror reported trends that psychosocial factors, revision status, and lower\hyp airway/systemic comorbidities blunt PROM gains.

\paragraph{Prediction and classification.}
The adjusted improvement was subtracted from baseline to obtain a predicted 6\hyp month SNOT\hyp 22 (clipped at 0); the difference \(\Delta=\) baseline \(-\) predicted was compared to MCID: \(\Delta>9 \Rightarrow 1\) (recommend), else \(0\).

\paragraph{Confidence mapping.}
Confidence used distance to the MCID threshold: \(|\Delta-9|\ge15\Rightarrow\) very confident; 10--14.9 \(\Rightarrow\) somewhat confident; 6--9.9 \(\Rightarrow\) neutral; 3--5.9 \(\Rightarrow\) somewhat unsure; \(<3 \Rightarrow\) not at all confident.

\paragraph{Assessment.}
Interestingly, after discussing this decision logic with the expert clinicians, we observed that this emergent heuristic aligns with clinician reasoning (baseline SNOT\hyp 22, CT/endoscopy severity, polyp phenotype, depression/fibromyalgia). Strengths include transparent dependence on symptom burden and objective disease and an explicit MCID\hyp anchored threshold. Limitations include (a) \emph{multiplicative stacking} that can over\hyp penalize co\hyp morbidities, (b) coarse confidence bins that yield poor probability calibration, and (c) sensitivity to bracket cutpoints chosen without data\hyp driven tuning. These factors help explain the LLM’s lower AUROC/AP versus the MLP, despite the qualitative alignment of its rationale with clinical heuristics.

\begin{figure}[t]
\centering
\includegraphics[width=\columnwidth]{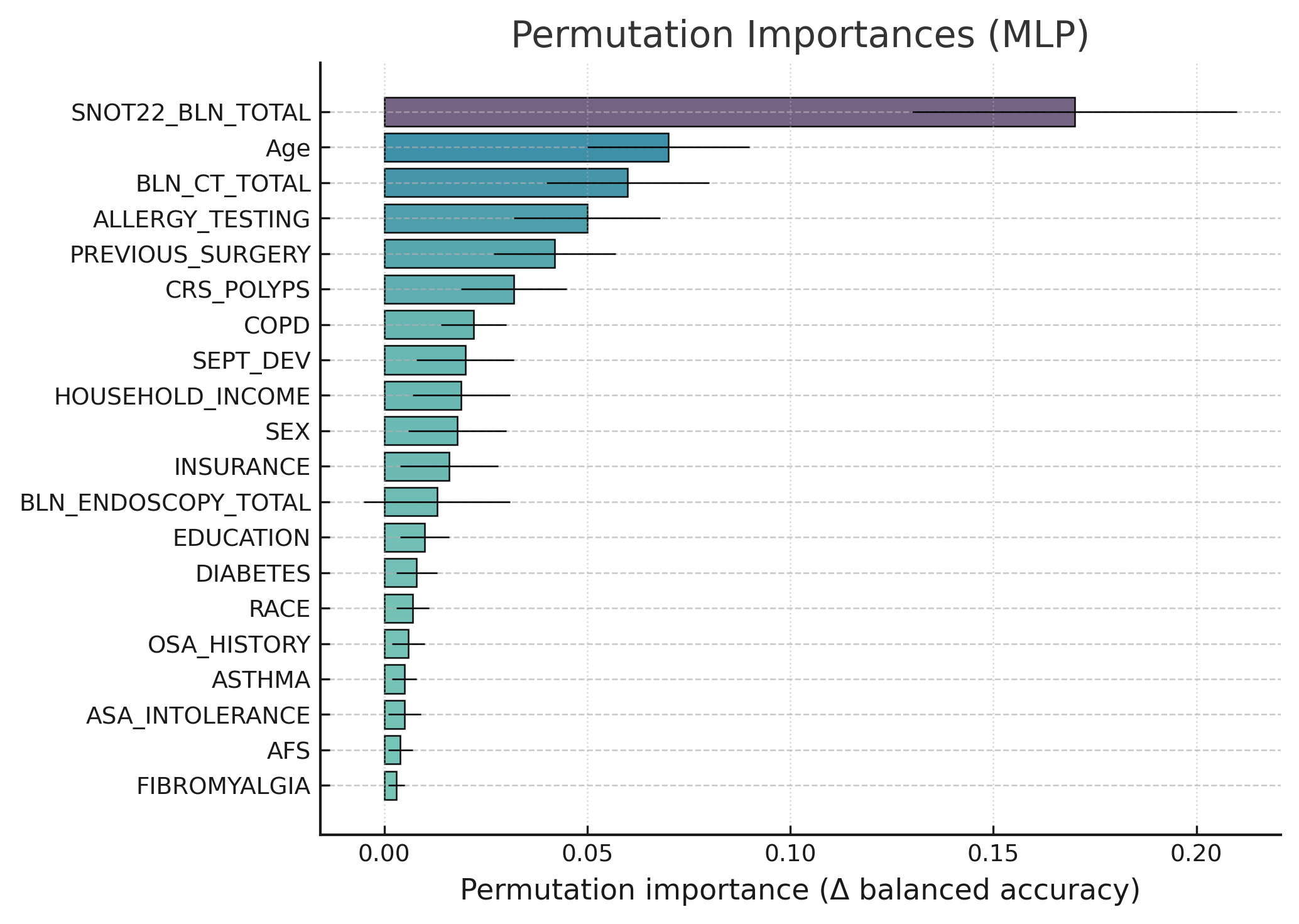}
\caption{Permutation feature importance for the MLP classifier on the held-out test set. Bars show the mean decrease in balanced accuracy when each feature is randomly permuted (larger values indicate greater importance); black ticks denote the standard deviation across permutation repeats. The model is most sensitive to \texttt{SNOT22\_BLN\_TOTAL}, \texttt{Age}, and \texttt{BLN\_CT\_TOTAL}, followed by \texttt{ALLERGY\_TESTING}, \texttt{PREVIOUS\_SURGERY}, and \texttt{CRS\_POLYPS}.}
\label{fig:mlp_feature_importance}
\end{figure}

\subsection{Feature importance analysis}

To make the model’s behavior inspectable, we computed \emph{permutation importance} on the held-out set, measuring the change in balanced accuracy ($\Delta$BA) when shuffling one feature at a time (means with SD across repeats; Fig.~\ref{fig:mlp_feature_importance}). This assesses the model’s dependence on each input without linearity assumptions or scaling artifacts.

\textit{What matters most.} The baseline SNOT-22 dominated as its permutation caused by far the largest drop in performance, followed by age and CT Lund–Mackay burden. Allergy and prior surgery also ranked highly, with additional contributions from polyp phenotype, COPD, and septal deviation. Socioeconomic and demographic fields (income, insurance, sex, race) showed smaller but nonzero effects and should be monitored for fairness rather than interpreted causally.

\textit{Convergent reasoning.} Notably, these top drivers mirror the factors emphasized by the LLM’s self-described decision rules (baseline symptoms, objective disease: CT/endoscopy, polyp status, and comorbidities such as depression/airway disease), and they align with clinician heuristics and guidelines \cite{c1,aao_indicators_ess,aao_guideline_adultsinusitis_2015,icar_rs_2021}. Thus, despite lower discrimination from GenAI, the \emph{qualitative} signals it highlights are consistent with both expert judgment and the MLP’s learned importance profile, supporting an ML-first, GenAI-augmented workflow where the LLM explains decisions in familiar clinical terms.

\subsection{ChatGPT with Retrieval Augmented Generation (RAG)}

We tested whether \emph{retrieval-augmented generation} (RAG) \cite{lewis2020rag} could strengthen ChatGPT’s predictions by grounding them in explicit CRS guidance. We assembled a compact corpus (EPOS/AAO-HNS excerpts, SNOT-22/MCID meta-analyses, brief CT/endoscopy summaries), indexed it (BM25), and prepended the top-$k{=}5$ passages to the serialized pre-operative features in the same constrained prompt (binary output + confidence). No labels were exposed; the identical test split and evaluation were used.

\begin{figure}[t]
  \centering
  \includegraphics[width=\columnwidth]{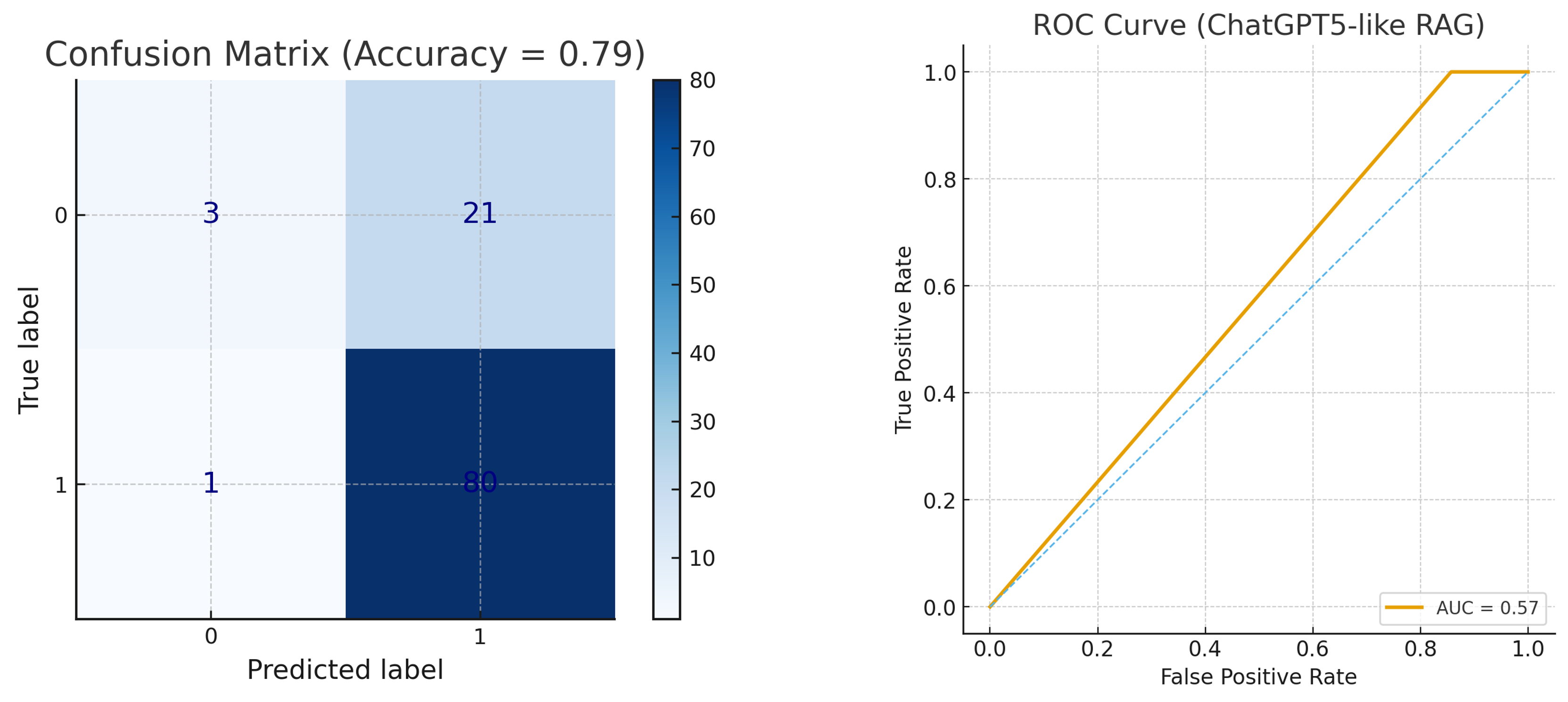}
  \caption{ChatGPT (GPT-5 Thinking) with RAG. Left: confusion matrix (TN$=3$, FP$=21$, FN$=1$, TP$=80$; accuracy $=0.79$). Right: ROC curve (AUC $=0.57$). RAG does not improve minority-class detection and leaves overall discrimination essentially unchanged.}
  \label{fig:chatgpt_rag}
\end{figure}

Figure~\ref{fig:chatgpt_rag} shows the outcome. With RAG, ChatGPT achieved accuracy $0.79$ and ROC–AUC $0.57$ with confusion matrix $[3,21;\,1,80]$ (TN,FP;FN,TP). Thus, Class~1 recall remained very high ($80/81\approx0.99$), but Class~0 recall was low ($3/24=0.125$), yielding \emph{21 false-positive} recommendations. Compared with the non-RAG setting (AUROC $0.58$; Sec.~\ref{fig:big_results_grid}), retrieval produced no meaningful improvement in discrimination or minority-class detection.

Why the limited effect? Our task is patient-specific \emph{prediction}, not fact recall. Retrieved guideline snippets reiterate broad heuristics (e.g., higher baseline SNOT-22 and objective burden predict larger gains) already internalized during pretraining, but they do not contain additional, individualized signal beyond the tabular inputs. Moreover, few passages map directly to our MCID decision threshold. In short, for structured, preoperative prediction, calibrated tabular ML (our MLP) remains superior, while GenAI, with or without RAG, is best positioned as an explanatory layer rather than the primary classifier.

\section{Conclusions}

We asked a clinically pointed question: using only pre-operative data, who \emph{should not} have been recommended ESS despite guideline eligibility? Across a fair head-to-head benchmark, a calibrated tabular model (our MLP) provided the strongest performance and clinical utility: highest AUROC, competitive F1 and the best minority-class (Class~0) detection—while multiple state-of-the-art LLMs performed poorly in discrimination and calibration. Yet, when probed, LLM rationales emphasized the same high-value signals that clinicians and the MLP prioritize (baseline SNOT-22, CT/endoscopy severity, polyp phenotype, mood/pain comorbidities), suggesting that GenAI can \emph{explain} decisions in clinically familiar terms even when it is not the best \emph{predictor}. Retrieval augmentation (RAG) did not close the performance gap, likely because guideline text reiterates general heuristics already internalized by LLMs and adds little patient-specific signal beyond structured features.

\paragraph{Implications for practice.}
These findings support an \textbf{ML-first, GenAI-augmented} workflow: deploy a calibrated tabular model for primary triage and individualized risk estimates (with reliability plots and decision-curve net benefit), and use GenAI to (i) translate the prediction into plain-language rationales tied to guideline concepts, (ii) surface patient-facing summaries for shared decision-making, and (iii) document auditable explanations. This division of labor aligns strengths with tasks: tabular ML for calibrated probabilities and thresholding; GenAI for communication, sense-making, and hypothesis generation.

\paragraph{Limitations.}
This study is retrospective, single clinical domain, and uses MCID $\geq 8.9$ as a pragmatic endpoint; labels contain inherent noise around that threshold. LLM outputs were constrained to binary/confidence schemas; alternate tool use (e.g., structured function-calling with calibrated heads) might improve reliability. Our RAG corpus was deliberately compact; broader corpora may aid other tasks (e.g., counseling, consent) more than numeric prediction.

\paragraph{Where to next.}
We see several concrete directions:
\begin{itemize}
  \item \textbf{Prospective, multi-site validation} with external cohorts, including CRSwNP/CRSsNP strata and low-baseline SNOT-22 patients; report decision-curve net benefit and \emph{cost-effectiveness} of alternative thresholds.
  \item \textbf{Selective prediction and abstention}: allow the classifier to defer when uncertainty is high; route such cases to MDT review with GenAI-assisted narratives.
  \item \textbf{Hybrid modeling}: couple a calibrated tabular backbone to an LLM explainer via tool-use APIs; study calibration-preserving post-hoc explanations and counterfactual recourse (``what would change the recommendation?'').
  \item \textbf{Governance and fairness}: continuous drift monitoring, subgroup calibration, and bias audits (socioeconomic/demographic variables) with mitigation policies; privacy-preserving training (federated learning) for wider deployment.
  \item \textbf{Workflow integration}: EHR connectors for automated feature extraction, patient-facing summaries in after-visit notes, and clinician-in-the-loop feedback to adapt thresholds to local preferences.
\end{itemize}

\noindent In sum, calibrated tabular ML currently offers the most dependable foundation for pre-operative ESS decision support, while GenAI adds value as an explanatory and engagement layer. Combining these strengths, rather than substituting one for the other offers a practical path to safer, more transparent, and patient-centered surgical decision-making in CRS and beyond.

\addtolength{\textheight}{-12cm}   % This command serves to balance the column lengths

%%%%%%%%%%%%%%%%%%%%%%%%%%%%%%%%%%%%%%%%%%%%%%%%%%%%%%%%%%%%%%%%%%%%%%%%%%%%%%%%

%%%%%%%%%%%%%%%%%%%%%%%%%%%%%%%%%%%%%%%%%%%%%%%%%%%%%%%%%%%%%%%%%%%%%%%%%%%%%%%%



\section*{ACKNOWLEDGMENT}

This work was supported by a grant from the National Institute of Allergy and Infectious Diseases (NIAID) of the National Institutes of Health under grant number R01AI175631 (VRR). Contents are the authors’ sole responsibility and do not necessarily represent official NIH views.

\end{document}